\documentclass[11pt]{article}
\pdfoutput=1
\usepackage{geometry}
\usepackage{natbib}
\setcitestyle{authoryear,open={(},close={)}} 

\usepackage{mathtools}
\usepackage{xcolor}
\newcommand{\bftab}{\fontseries{b}\selectfont}
\usepackage{algpseudocode}
\usepackage{subcaption}
\usepackage{makecell}
\usepackage[pdftex,bookmarksnumbered,bookmarksopen,
colorlinks,citecolor=blue,linkcolor=blue,urlcolor=blue]{hyperref}
\usepackage{algorithm}

\title{Robust Optimization as Data Augmentation for Large-scale Graphs\footnote{Accepted at CVPR 2022.}}

\author{
Kezhi Kong \\ University of Maryland \\ \small{kong@cs.umd.edu} \and
Guohao Li \\ KAUST \\ \small{guohao.li@kaust.edu.sa} \and
Mucong Ding \\ University of Maryland \\ \small{mcding@cs.umd.edu} \and
Zuxuan Wu \\ University of Maryland \\ \small{zxwu@cs.umd.edu} \and
Chen Zhu \\ University of Maryland \\ \small{chenzhu@cs.umd.edu}  \and
Bernard Ghanem \\ KAUST \\ \small{bernard.ghanem@kaust.edu.sa} \and 
Gavin Taylor \\ US Naval Academy \\ \small{taylor@usna.edu} \and 
Tom Goldstein \\  University of Maryland \\ \small{tomg@cs.umd.edu}
}

\date{}

\begin{document}

\maketitle

\begin{abstract}
Data augmentation helps neural networks generalize better by enlarging the training set, but it remains an open question how to effectively augment graph data to enhance the performance of GNNs (Graph Neural Networks). While most existing graph regularizers focus on manipulating graph topological structures by adding/removing edges, we offer a method to augment node features for better performance. We propose FLAG (Free Large-scale Adversarial Augmentation on Graphs), which iteratively augments node features with gradient-based adversarial perturbations during training. By making the model invariant to small fluctuations in input data, our method helps models generalize to out-of-distribution samples and boosts model performance at test time. FLAG is a general-purpose approach for graph data, which universally works in node classification, link prediction, and graph classification tasks. FLAG is also highly flexible and scalable, and is deployable with arbitrary GNN backbones and large-scale datasets. We demonstrate the efficacy and stability of our method through extensive experiments and ablation studies. We also provide intuitive observations for a deeper understanding of our method. We open source our implementation at \url{https://github.com/devnkong/FLAG}.
\end{abstract}

\section{Introduction}


Graph Neural Networks (GNNs) have emerged as powerful architectures for learning and analyzing graph representations. The Graph Convolutional Network (GCN) \citep{kipf2016semi} and its variants have been applied to a wide range of tasks, including visual recognition \citep{shen2018person}, meta-learning \citep{garcia2017few}, social analysis \citep{qiu2018deepinf,li2019encoding}, and recommender systems \citep{ying2018graph}. However, the training of GNNs on large-scale datasets usually suffers from overfitting, and realistic graph datasets often involve a high volume of out-of-distribution test nodes \citep{hu2020open}, posing significant challenges for prediction problems. 

One promising solution to combat overfitting in deep neural networks is data augmentation \citep{krizhevsky2012imagenet}, which is commonplace in computer vision tasks. Data augmentations apply label-preserving transformations to the inputs, such as translations and reflections for images. As a result, data augmentation effectively enlarges the training set while incurring negligible computational overhead. However, it remains an open problem how to effectively generalize the notion of data augmentation to GNNs. Transformations on images rely heavily on image structures \citep{chen2020simple}, and it is challenging to design low-cost transformations that preserve semantic meaning for non-visual tasks like natural language processing \citep{wei2019eda} and graph learning. Generally speaking, graph data for machine learning comes with graph structure (or edge features) and node features. In the limited cases where data augmentation can be done on graphs, it generally focuses exclusively on the graph structure by adding/removing edges \citep{rong2019dropedge,hamilton2017inductive,gilmer2017neural,you2020graph,wang2020nodeaug,godwin2022simple}. 

 
\begin{figure*}[t]%
    \centering
    \begin{subfigure}{.3\textwidth}
        \includegraphics[width=\linewidth]{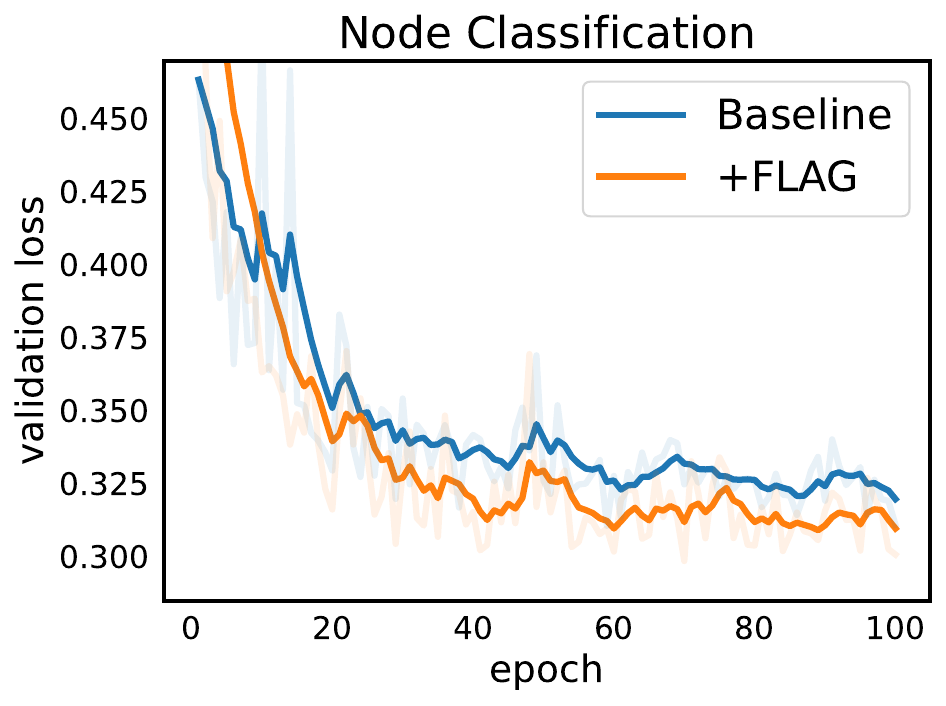}%
    \end{subfigure}
    \begin{subfigure}{.3\textwidth}
        \includegraphics[width=\linewidth]{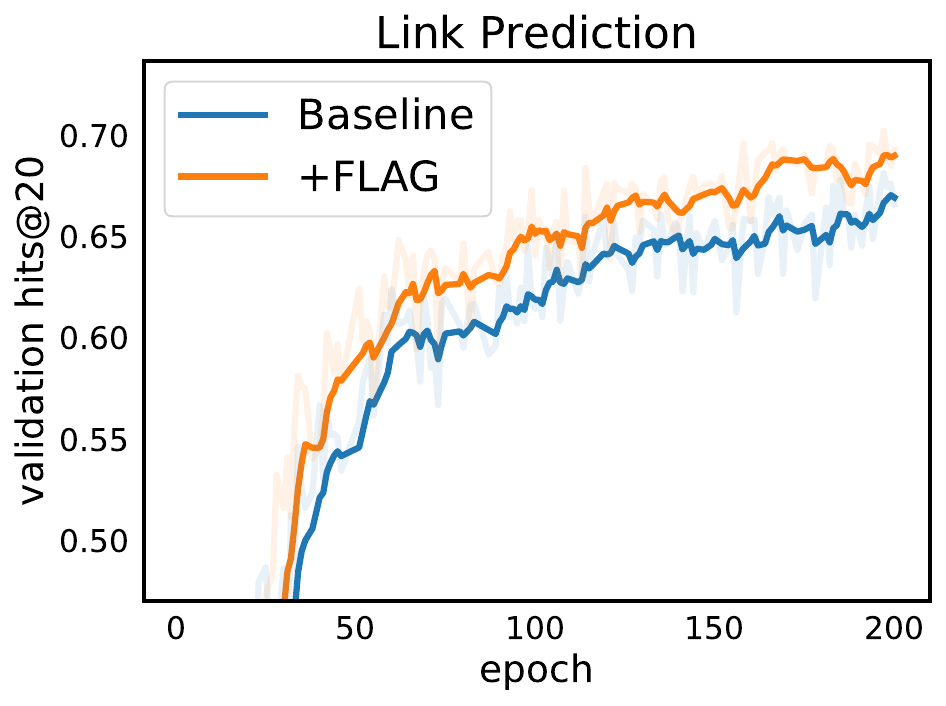}%
    \end{subfigure}
    \begin{subfigure}{.3\textwidth}
        \includegraphics[width=\linewidth]{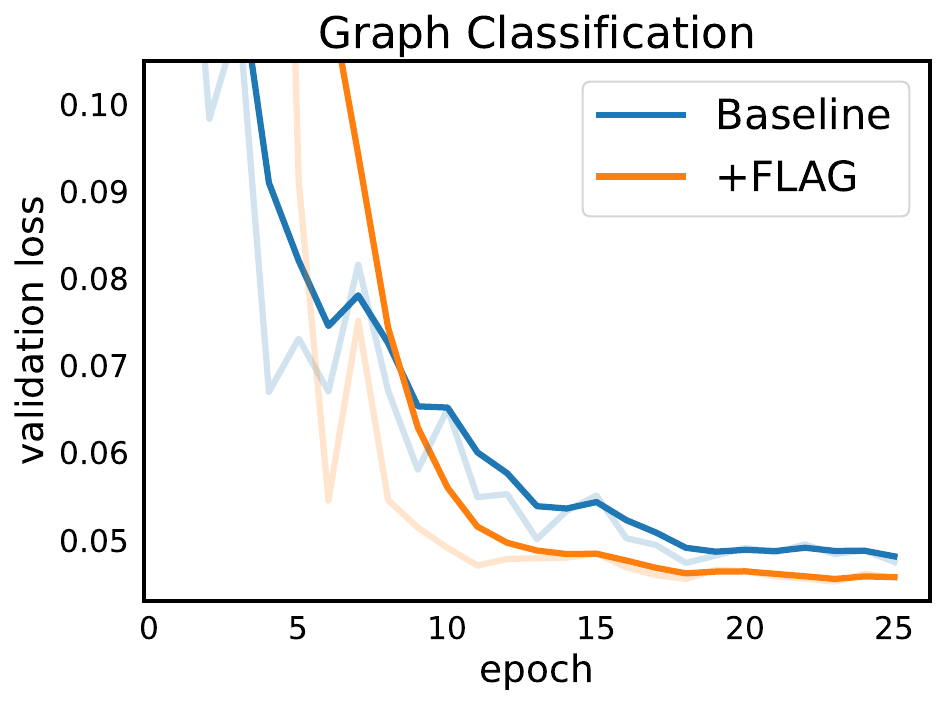}
    \end{subfigure}
    \caption{Generalization performance of FLAG on all three tasks. Left: node classification with GAT as baseline on \texttt{ogbn-products}; Middle: link prediction with hits@20 as metric (the higher the better) and GraphSAGE as baseline on \texttt{ogbl-ddi}; Right: graph classification with GIN as baseline on \texttt{ogbg-molhiv}. Plotted lines are attained by smoothing the original lines (the shallow ones), where smooth weights are 0.75, 0.75, and 0.5 respectively.}%
    \label{fig:loss}%
\end{figure*}

In the meantime, adversarial data augmentation, which applies small perturbations in the input feature space to maximially alter model outputs, is known to boost neural network robustness and promote resistance to adversarially chosen inputs \citep{goodfellow2014explaining,madry2017towards}. Despite the wide belief that adversarial training harms standard generalization and leads to worse accuracy \citep{tsipras2018robustness,balaji2019instance}, recently a growing amount of attention has been paid to using adversarial perturbations to augment datasets and ultimately alleviate overfitting. For example, \citet{volpi2018generalizing} and \citet{shu2020prepare} showed adversarial data augmentation is a data-dependent regularization that could help generalize to out-of-distribution samples, and its efficacy has been verified in domains including computer vision \citep{xie2020adversarial}, language understanding \citep{miyato2016adversarial,zhu2019freelb, jiang2019smart}, and visual question answering \citep{gan2020large}. Despite the success of adversarial augmentation in language and vision, it remains unclear how to effectively and efficiently improve GNNs' clean accuracy using adversarial augmentation. 

{\bf Present work.} We propose {\bf FLAG}, \textbf{F}ree \textbf{L}arge-scale \textbf{A}dversarial Augmentation on \textbf{G}raphs, to tackle the overfitting problem. While existing literature focuses on modifying graph structures to augment datasets, FLAG works purely in the node feature space by adding  adversarial perturbations (generated by gradient-based robust optimization algorithms), to the input node features with graph structures unchanged. FLAG leverages ``free" adversarial training methods \citep{shafahi2019adversarial} to conduct efficient adversarial training so that it is highly scalable to large datasets. The method also takes advantage of multi-scale adversarial augmentation to make the model fully generalized in the input feature space. We verify the effectiveness of our method on the \textit{Open Graph Benchmark} (OGB) \citep{hu2020open}, which is a collection of large-scale, realistic, and diverse graph datasets for node, link, and graph property prediction tasks. We conduct extensive experiments across OGB datasets by applying FLAG to competitive GNN baselines and show that FLAG brings non-trivial improvements in most cases. For example, FLAG lifts the test accuracy of GAT on \texttt{ogbn-products} by an absolute value of 2.31\%. FLAG is simple (easy to implement with a dozen lines of code in PyTorch), general (model-free and task-free), and efficient (able to bring salient improvement at tractable or even no extra cost). Our main contributions are summarized as follows: 

\begin{itemize}
    \item \textit{Method:} To the best of our knowledge, our work is the \textit{first} general-purpose feature-based data augmentation method on graph data, which is complementary to other regularizers (e.g., dropout) and topological augmentations. The novel method incorporates ``free'' and multi-scale techniques to craft feature augmentations more effectively.
  \item \textit{Experiments:} We show the efficacy and scalability of our method through extensive experiments and ablation studies on large-scale datasets across node, link, and graph property prediction tasks. We validate that FLAG is superior to existing adversarial augmentation methods.
  \item \textit{Analysis:} We provide observations and analysis to support our conjecture that the discrete vs. continuous distribution discrepancy of input features is the key to different effects (beneficial vs. harmful) of adversarial augmentations on model accuracy.
\end{itemize}

\section{Preliminaries and Related Work}\label{sec:pre}

{\bf Graph Neural Networks (GNNs).} We denote a graph as $\mathcal{G}(\mathcal{V}, \mathcal{E})$ with initial node features $x_{v}$ for $v \in \mathcal{V}$ and edge features $e_{u v}$ for $(u, v) \in \mathcal{E}$. GNNs are built on graph structures to learn representation vectors $h_{v}$ for every node $v \in \mathcal{V}$ and a vector $h_{\mathcal{G}}$ for the entire graph $\mathcal{G}$. Following \citet{hu2019strategies}, formally the $k$-th iteration of message passing, or the $k$-th layer of GNN forward path is defined as:
\begin{equation}
\label{eqn:mp}
\begin{aligned}
        msg_{v}^{(k)} & = \text{AGGREGATE}^{(k)}_{\theta}\left( \left\{\left(h_{v}^{(k-1)}, h_{u}^{(k-1)}, e_{u v}\right), \forall u \in \mathcal{N}(v)\right\}\right) \\
        h_{v}^{(k)} & = \text{COMBINE}^{(k)}_{\phi}\left(h_{v}^{(k-1)}, msg_{v}^{(k)}\right),
\end{aligned}
\end{equation}
where $h_{v}^{(k)}$ is the embedding of node $v$ at the $k$-th layer, $e_{uv}$ is the feature vector of the edge between node $u$ and $v$, $\mathcal{N}(v)$ is node $v$'s neighbor set, and $h_{v}^{(0)}=x_{v}$. AGGREGATE($\cdot$) and COMBINE($\cdot$) functions are parameterized by neural networks. 


To obtain the representation of the entire graph $h_\mathcal{G}$, the permutation-invariant READOUT($\cdot$) function pools node features from the final iteration $K$ as:
\noindent
\begin{equation}
\label{eqn:readout}
 h_{\mathcal{G}}=\text{READOUT}\left(\left\{ h_{v}^{(K)} \mid v \in \mathcal{V}\right\}\right),
\end{equation}
\noindent

Existing graph regularizers mainly focus on augmenting graph structures by modifying edges \citep{rong2019dropedge,hamilton2017inductive,chen2018fastgcn}. GraphAT \citep{feng2019graph}, BVAT \citep{deng2019batch}, and LAT \citep{jin2019latent} are three \textit{semi-supervised} methods on the node classification task. GraphAT promotes local smoothness by reinforcing the similarity between the predictions of perturbed nodes and their neighbors. BVAT proposed two graph VAT schemes to enhance the output smoothness of GCN; LAT virtually perturbed the first-layer embedding of a GCN classifier. The usage scenario of these methods is limited to node classification, while data augmentation should function regardless of tasks. Besides, the formulation of VAT \citep{miyato2018virtual} utilized by these works involves both supervised clean and adversarial robust losses simultaneously. Practically this will consume at least twice the GPU memory as the baseline, making them not scalable to large-scale datasets. Overall, no work so far has considered general-purpose feature-based data augmentations for large-scale graphs.

\section{Proposed Method}

In this work, we investigate how to effectively improve the generalization of GNNs through a feature-based augmentation. Graph node features are usually constructed as discrete embeddings, such as binary bag-of-words vectors or categorical variables.  As a result, standard hand-crafted augmentations, like flipping and cropping transforms used in computer vision, are not applicable to graphs node features.   


By hunting for and stamping out small perturbations that cause the classifier to fail, one may hope that adversarial training could benefit standard accuracy \citep{goodfellow2014explaining,tsipras2018robustness,miyato2018virtual}. It is widely observed that when the data distribution is sparse and discrete, the beneficial effect of adversarial perturbations on generalization takes over \citep{tsipras2018robustness,gan2020large}. \citet{volpi2018generalizing} viewed adversarial perturbation as a data-dependent regularization, which could intuitively generalize to out-of-distribution samples. Highlighted by \citet{hu2020open}, the out-of-distribution phenomenon of data is salient in the graph domain, and also considering the sparsity of labeled node samples in the semi-supervised node classification task, we view adversarial perturbation as a strong candidate method for input feature augmentation.

{\bf Min-Max Optimization.} Adversarial training is the process of crafting adversarial data points, and then injecting them intro training data. This process is often formulated as the following min-max problem:
\begin{equation}
\min _{\boldsymbol{\theta}}~~ {E}_{(x,y) \sim \mathcal{D}}\left[\max _{\|\boldsymbol{\delta}\|_p \leq \epsilon} L\left(f_{\boldsymbol{\theta}}( x+\boldsymbol{\delta}), y\right)\right],\label{eq:minmax}
\end{equation}
where $\mathcal{D}$ is the data distribution, $y$ is the label, $\|\cdot\|_{p}$ is some $\ell_{p}$-norm distance metric, $\epsilon$ is the perturbation budget, and $L$ is the objective function. \citet{madry2017towards} showed that this saddle-point optimization problem could be reliably tackled by Stochastic Gradient Descent (SGD) for the outer minimization and Projected Gradient Descent (PGD) for the inner maximization. In practice, the typical approximation of the inner maximization under an $l_\infty$-norm constraint is as follows,
\begin{equation}
\boldsymbol{\delta}_{t+1}=\Pi_{\|\boldsymbol{\delta}\|_{\infty} \leq \epsilon}\left(\boldsymbol{\delta}_{t}+\alpha \cdot \text{sign} \left( \nabla_{\boldsymbol{\delta}} L\left(f_{\boldsymbol{\theta}}( x+\boldsymbol{\delta}_t), y\right) \right) \right),\label{eq:pgd}
\end{equation}
where the perturbation $\boldsymbol{\delta}$ is updated iteratively, and $\Pi_{\|\boldsymbol{\delta}\|_{\infty} \leq \epsilon}$ performs projection onto the $\epsilon$-ball in the $l_{\infty}$-norm. For maximum robustness, this iterative updating procedure usually loops $M$ times to craft the worst-case noise, which requires $M$ forward and backward passes end-to-end. Afterwards the most vicious noise $\boldsymbol{\delta}_M$ is applied to the input feature, on which the model weight is optimized. The algorithm above is called PGD.

\textbf{Multi-scale Augmentation.} On visual tasks, \citet{chen2020simple} highlighted the importance of using \textit{diverse} types of data augmentations such as \textit{random cropping}, \textit{color distortion}, and \textit{Gaussian blur}. The authors showed that a single transformation is not sufficient to learn good representations. To fully exploit the generalizing ability and enhance the diversity and quality of adversarial perturbations, we propose to craft multi-scale augmentations. To realize this goal, we leverage the techniques below.

\begin{figure}[t]%
    \centering
    \includegraphics[width=0.5\linewidth]{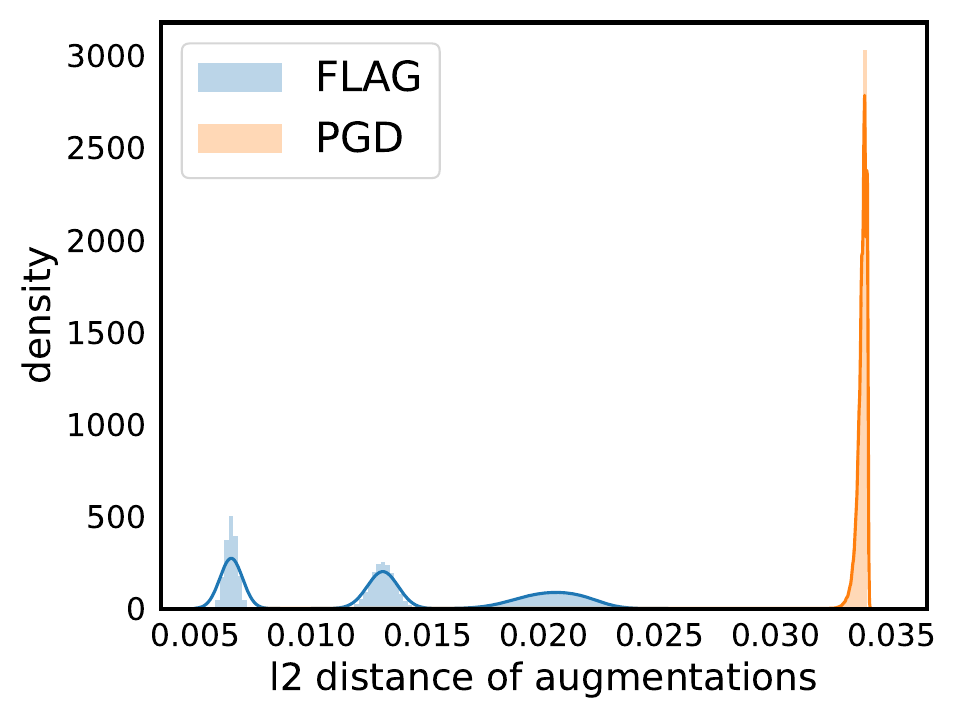} %
    \caption{Augmentation distance distributions of FLAG and PGD. We run the test on \texttt{ogbn-arxiv} with GCN as backbone. Ascent steps are both set to 3.}%
    \label{fig:hist}%
\end{figure}

\textit{``Free'' training.} We leverage ``free'' adversarial training \citep{shafahi2019adversarial} to craft adversarial data augmentations. PGD is a powerful yet inefficient way of solving the min-max optimization. It runs $M$ full forward and backward passes to craft a refined perturbation $\boldsymbol{\delta}_{1:M}$, but the model weights $\boldsymbol{\theta}$ only get updated once using the final $\boldsymbol{\delta}_M$.  This process makes model training $M$ times slower. In contrast, while computing the gradient for the perturbation $\boldsymbol{\delta}$, ``free" training simultaneously produces the model parameter $\boldsymbol{\theta}$ on the same backward pass. This enables a parameter update to be computed in parallel with a perturbation update at virtually no additional cost. The authors proposed to train on the same minibatch $M$ times in a row to simulate the inner maximization in Eq. (\ref{eq:minmax}), while compensating by performing $M$ times fewer epochs of training. The resulting algorithm yields accuracy and robustness competitive with standard adversarial training, but with the same runtime as clean training.

Besides the efficiency, the ``free'' method achieves our idea of optimizing $\boldsymbol{\theta}$ with multi-scale augmentations. Note that $X$ is augmented with additive perturbations $\boldsymbol{\delta}_{1:M}$, of which each can have a maximum scale of $m\alpha, m \in \left\{1, \cdots,M\right\}$, in contrast to PGD whose perturbation is a single $\boldsymbol{\delta}_M$ with an $M\alpha$ scaling. This greatly adds to the diversity of our augmentations. However, the ``free'' algorithm is suboptimal in terms of min-max optimization in that during the batch-replay process, the approximated perturbation computed to maximize the objective on $\boldsymbol{\theta_t}$ is used to robustly optimize $\boldsymbol{\theta_{t+1}}$ rather than $\boldsymbol{\theta_t}$. To tackle this problem, instead of directly updating $\boldsymbol{\theta}$ using the ``by-product'' gradient attained from the gradient ascent step on $\boldsymbol{\delta}$, we accumulate the gradients $\nabla_{\boldsymbol{\theta}} L$, and apply them to the model parameters all at once later. Formally, the optimization step is 
\begin{equation}\label{eq:update}
    \boldsymbol{\theta}_{i+1} = \boldsymbol{\theta}_i - \frac{\tau}{M}\sum_{t=1}^{M} \nabla_{\boldsymbol{\theta}}  L \left( f_{\boldsymbol{\theta}}( x+\boldsymbol{\delta_t}), y \right),
\end{equation}
where $\tau$ is learning rate and $\boldsymbol{\delta}_1$ is uniform noise. Note that the gradients in Eq.(\ref{eq:update}) are restored when crafting perturbation in Eq.(\ref{eq:pgd}). We save one backward pass and $M$ times extra GPU memory through accumulating gradients (which is fully supported by PyTorch) during the batch replay process. Figure \ref{fig:hist} depicts the effects of our design. We can see that PGD inevitably produces concentrated augmentations in terms of the magnitude, whereas our method produces perturbations with a broader range of sizes, which adds to the diversity and quality of the augmentations.

Moreover on the node classification task, we propose to augment labeled vs. unlabeled nodes with diverse magnitudes of perturbations during training time to further diversify the augmentations. We call it \textit{Weighted perturbation}. When classifying one target node, messages from the whole $k$-hop neighborhood are aggregated and combined into its embedding. It is natural to believe that a further neighbor should have lower impact, i.e. higher smoothness, on the final decision of the target node, which can also be intuitively reflected by the recursive message passing procedure of GNNs in Eq.(\ref{eqn:mp}). In practice we find that a larger perturbation for unlabeled nodes can be beneficial to the performance.  Algorithm~\ref{alg:node} summarizes the pseudo code of our method on node classification task. Figure~\ref{fig:loss} illustrates the generalization ability of our proposed method.

\begin{algorithm}[tb] 
  \caption{\textbf{FLAG}: \textbf{F}ree \textbf{L}arge-scale \textbf{A}dversarial Augmentation on \textbf{G}raphs (Node Classification Task)}
  \label{alg:node}
  \textbf{Require:} Graph $\mathcal{G} = (\mathcal{V},\mathcal{E})$, $\mathcal{V}_l$ is the labeled node set; learning rate $\tau$; ascent steps $M$; ascent step size $\alpha_v$ for labeled node, $\alpha_u$ for unlabeled, we assume the neighbors of labeled nodes are all unlabeled ones; $L(\cdot)$ as objective function; A($\cdot$) and C($\cdot$) denote the AGGREGATE and COMBINE functions in Eq.(\ref{eqn:mp}). The backward function at line 12 refers to back-propagation gradient computation for both model weights and noises.
  \begin{algorithmic}[1] 
    \State Initialize $(\theta,\phi)$
        
        \For{$v \in \mathcal{V}_l$}

            \State $\delta_{v}^{(0)} \leftarrow  U(-\alpha_v, \alpha_v)$ 
            \State $\delta_{u}^{(0)} \leftarrow  U(-\alpha_u, \alpha_u)$ 
            \For{$\text{t} = 1 \ldots M$}
                \State $h_{v}^{(0)} \leftarrow x_{v} + \delta_{v}^{(t-1)}$
                \State $h_{u}^{(0)} \leftarrow x_{u} + \delta_{u}^{(t-1)}$
                \For{$k = 1 \ldots K$}
                    \State $msg_{v}^{(k)} \leftarrow \text{A}^{(k)}_{\theta}\left( \left\{\left(h_{v}^{(k-1)}, h_{u}^{(k-1)}, e_{u v}\right), \forall u \in \mathcal{N}(v)\right\}\right)$
                    \State $h_{v}^{(k)}  \leftarrow \text{C}^{(k)}_{\phi}\left(h_{v}^{(k-1)}, msg_{v}^{(k)}\right)$
                \EndFor 
                
                \State $L\left(h_{v}^{(K)}, y\right).\text{backward}()$
                \State $g_{\theta,\phi}^{(t)} \leftarrow g_{\theta,\phi}^{(t-1)}+\frac{1}{M} \cdot \text{grad}(\theta,\phi)$
                
                
                \State $\delta_{v}^{(t)} \leftarrow \delta_{v}^{(t-1)}+\alpha_{v} \cdot \text{sign} \left(\text{grad}\left(\delta_v\right)\right) $
                
                \State $\delta_{u}^{(t)} \leftarrow \delta_{u}^{(t-1)}+\alpha_{u} \cdot \text{sign} \left(\text{grad}\left(\delta_u\right)\right) $
                
            \EndFor
            \State $(\theta,\phi) \leftarrow (\theta,\phi)-\tau \cdot g_{\theta,\phi}^{(M)}$
            
        \EndFor
    
  \end{algorithmic}
\end{algorithm}

\section{Experiments}
\label{exp}

In this section, we conduct extensive experiments to fully reveal the efficacy of our method.



{\bf Datasets.} We demonstrate FLAG's effectiveness through extensive experiments on the \textit{Open Graph Benchmark} (OGB), which consists of a wide range of challenging large-scale datasets. \citet{shchur2018pitfalls}, \citet{errica2019fair}, and \citet{dwivedi2020benchmarking} showed that traditional graph datasets suffered from problems such as unrealistic and arbitrary data splits, highly limited data sizes, non-rigorous evaluation metrics, and common neglect of validation set, etc. In order to empirically study FLAG's effects in a fair and reliable manner, we conduct experiments on the OGB \citep{hu2020open} datasets, which have tackled those major issues and brought more realistic challenges to the graph research community. 


{\bf Setup.}  FLAG drops the projection step when performing the inner maximization, in light of the positive effect of large perturbations on generalization \citep{volpi2018generalizing}, and also to simplify hyperparameter search. Usually on images, the inner maximization has a norm constraint on the perturbation; the largest perturbation one can add is bounded by the hyperparameter $\epsilon$, typically 8/255 under the $l_{\infty}$-norm. This $\epsilon$ encourages the visual imperceptibility of the perturbations, thus making defenses realistic and practical. However, graph node features or language word embeddings do not have an established distance threshold for imperceptibility, which makes the selection of $\epsilon$ highly heuristic. Note that, although the perturbation is no longer bounded by an explicit $\epsilon$ in FLAG, it is still implicitly bounded in the furthest distance that $\boldsymbol{\delta}$ can reach, i.e. the step size $\alpha$ times the number of ascending steps $M$. 

Also unless otherwise stated, all of the baseline test statistics come from the official OGB leaderboard website, and we conduct all of our experiments using publicly released implementations without touching the original model architecture or training setup for fair comparisons. We report mean and standard deviations from 10 runs with different random seeds. Following common practice on this benchmark, we report the test performance associated with the best validation result. We choose GCN, GraphSAGE, GAT, and GIN as our baseline models. In addition, we apply FLAG to the DeeperGCN model to demonstrate its effectiveness on the GNNs with significantly deeper depth. Our implementation always uses $M=3$ ascent steps for simplicity. Following \citet{goodfellow2014explaining,madry2017towards}, we use sign$(\cdot)$ for gradient normalization.

{\bf Large-scale Node Property Prediction.} We summarize the results of node classification in Table \ref{tab:node-main}. Notably, FLAG yields a 2.31\% test accuracy lift for GAT, making GAT competitive on the \texttt{ogbn-products} dataset. Considering the specialty of not having input node features in \texttt{ogbn-proteins}, we provide detailed discussions on the effect of different node feature constructions in Section \ref{sec:ablate}. \texttt{ogbn-mag} is a heterogeneous network where only ``paper" nodes come with node features. We use the neighbor sampling mini-batch algorithm to train R-GCN and report its results in the Table \ref{tab:mag}. Surprisingly, FLAG can also directly bring nontrivial accuracy improvement without special designs for heterogeneous graphs, which demonstrates its versatility.   

\begin{table}[t] 
	\caption{Node property prediction test performance on \texttt{ogbn-products}, \texttt{ogbn-proteins},  and \texttt{ogbn-arxiv} datasets. Blank denotes no statistics on the leaderboard.}
\centering
\begin{tabular}{lccc}
\Xhline{2\arrayrulewidth}
 	& \texttt{ogbn-products} & \texttt{ogbn-proteins} & \texttt{ogbn-arxiv}   \\
 Backbone	& Test Acc      & Test ROC-AUC & Test Acc \\
\hline\hline
GCN & - & \bftab72.51\textcolor{gray}{  {$\pm$0.35}}   			&  71.74\textcolor{gray}{  {$\pm$0.29}}               \\
+FLAG & -  & 71.71\textcolor{gray}{  {$\pm$0.50}}              			&  \bftab72.04\textcolor{gray}{  {$\pm$0.20}}     \\
\hline
GraphSAGE     & 78.70\textcolor{gray}{  {$\pm$0.36}} & \bftab77.68 \textcolor{gray}{  {$\pm$0.20}}  & 71.49\textcolor{gray}{  {$\pm$0.27}}     \\
+FLAG   & \bftab79.36\textcolor{gray}{  {$\pm$0.57}}  & 76.57\textcolor{gray}{  {$\pm$0.75}}  &\bftab72.19\textcolor{gray}{  {$\pm$0.21}}           \\
\hline
GAT                     & 79.45\textcolor{gray}{  {$\pm$0.59}}  &   -     &      73.65\textcolor{gray}{  {$\pm$0.11}}         \\
+FLAG                   & \bftab81.76\textcolor{gray}{  {$\pm$0.45}}  &  -   & \bftab73.71\textcolor{gray}{  {$\pm$0.13}}           \\
\hline
DeeperGCN               & 80.98\textcolor{gray}{  {$\pm$0.20}}     & 85.80\textcolor{gray}{  {$\pm$0.17}}    & 71.92\textcolor{gray}{  {$\pm$0.16}}   \\
+FLAG                   & \bftab81.93\textcolor{gray}{  {$\pm$0.31}}  &     \bftab85.96\textcolor{gray}{  {$\pm$0.27}}    &     \bftab72.14\textcolor{gray}{  {$\pm$0.19}}         \\
\Xhline{2\arrayrulewidth}
	\end{tabular}

	\label{tab:node-main}
\end{table}

\begin{table}[ht] 
\caption{Test performance on the heterogeneous OGB node property prediction dataset \texttt{ogbn-mag}.}
        \centering
        \begin{tabular}{lc}
            \Xhline{2\arrayrulewidth}
            & \texttt{ogbn-mag} \\
            Backbone	& Test Acc  \\
            \hline\hline
            R-GCN & 46.78\textcolor{gray}{   {$\pm$0.67}} \\
            +FLAG  & \bftab47.37\textcolor{gray}{   {$\pm$0.48}}\\
            \Xhline{2\arrayrulewidth}
        \end{tabular}
\label{tab:mag}
\end{table}
\begin{table}[ht] 
	\caption{Link property prediction test performance on \texttt{ogbl-ddi} and \texttt{ogbl-collab} datasets.}
\centering
\begin{tabular}{lcc}
\Xhline{2\arrayrulewidth}
 	       & \texttt{ogbl-ddi} & \texttt{ogbl-collab} \\
 Backbone	& Hits@20 & Hits@50  \\
\hline\hline
GCN &  37.07 \textcolor{gray}{    {$\pm$5.07}}   			&  44.75\textcolor{gray}{    {$\pm$1.07}}               \\
+FLAG & \bftab 51.41\textcolor{gray}{    {$\pm$3.76}}              			&  \bftab46.22\textcolor{gray}{    {$\pm$0.81}}     \\
\hline
GraphSAGE  & 53.90 \textcolor{gray}{    {$\pm$4.74}} & 48.10 \textcolor{gray}{    {$\pm$0.81}}    \\
+FLAG   & \bftab63.31\textcolor{gray}{    {$\pm$6.06}}  & \bftab 48.44\textcolor{gray}{    {$\pm$0.40}}    \\
\Xhline{2\arrayrulewidth}
	\end{tabular}

	\label{tab:link}
\end{table}



\begin{table}[t]
\caption{Graph property test performance on \texttt{ogbg-molhiv},  \texttt{ogbg-molpcba}, \texttt{ogbg-ppa}, and \texttt{ogbg-code} datasets. $\natural$ denotes the existence of virtual nodes; blank denotes no statistics on the leaderboard.}
\centering
\begin{tabular}{lcccc}
\Xhline{2\arrayrulewidth}
 	& \texttt{ogbg-molhiv} & \texttt{ogbg-molpcba} & \texttt{ogbg-ppa} & \texttt{ogbg-code}   \\
Backbone 	& Test ROC-AUC & Test AP & Test Acc & Test F1 \\
\hline\hline
GCN & 76.06\textcolor{gray}{{$\pm$0.97}}  & 20.20\textcolor{gray}{ {$\pm$0.24}}& 68.39\textcolor{gray}{ {$\pm$0.34}} &  31.63\textcolor{gray}{ {$\pm$0.18}}\\
+FLAG & \bftab76.83\textcolor{gray}{ {$\pm$1.02}} & \bftab21.16\textcolor{gray}{ {$\pm$0.17}}& 68.38\textcolor{gray}{ {$\pm$0.47}}& \bftab32.09\textcolor{gray}{ {$\pm$0.19}}\\
\hline
GCN-Virtual& \bftab75.99\textcolor{gray}{ {$\pm$1.19}} & 24.24\textcolor{gray}{ {$\pm$0.34}} & 68.57\textcolor{gray}{ {$\pm$0.61}}&32.63\textcolor{gray}{ {$\pm$0.13}} \\
+FLAG&75.45\textcolor{gray}{ {$\pm$1.58}} & \bftab 24.83\textcolor{gray}{ {$\pm$0.37}}& \bftab69.44\textcolor{gray}{ {$\pm$0.52}}&\bftab33.16\textcolor{gray}{ {$\pm$0.25}} \\
\hline
GIN& 75.58\textcolor{gray}{ {$\pm$1.40}} & 22.66\textcolor{gray}{ {$\pm$0.28}}&68.92\textcolor{gray}{ {$\pm$1.00}}&31.63\textcolor{gray}{ {$\pm$0.20}} \\
+FLAG& \bftab76.54\textcolor{gray}{ {$\pm$1.14}}&\bftab23.95\textcolor{gray}{ {$\pm$0.40}}&\bftab69.05\textcolor{gray}{ {$\pm$0.92}}&\bftab32.41\textcolor{gray}{ {$\pm$0.40}} \\
\hline
GIN-Virtual& 77.07\textcolor{gray}{ {$\pm$1.49}}& 27.03\textcolor{gray}{ {$\pm$0.23}}&70.37\textcolor{gray}{ {$\pm$1.07}}&32.04\textcolor{gray}{ {$\pm$0.18}} \\
+FLAG& \bftab77.48\textcolor{gray}{ {$\pm$0.96}} & \bftab28.34\textcolor{gray}{ {$\pm$0.38}}&\bftab72.45\textcolor{gray}{ {$\pm$1.14}}&\bftab32.96\textcolor{gray}{ {$\pm$0.36}} \\
\hline
DeeperGCN& 78.58\textcolor{gray}{ {$\pm$1.17}} & 27.81$^\natural$\textcolor{gray}{ {$\pm$0.38}} & 77.12\textcolor{gray}{ {$\pm$0.71}}&-\\
+FLAG& \bftab79.42\textcolor{gray}{ {$\pm$1.20}}&\bftab28.42$^\natural$\textcolor{gray}{ {$\pm$}0.43}&\bftab77.52\textcolor{gray}{ {$\pm$0.69}}&-\\
\Xhline{2\arrayrulewidth}

\end{tabular}

\label{tab:graph}
\end{table}

{\bf Large-scale Link Property Prediction.} We evaluate our method on two OGB link prediction datasets, which are \texttt{ogbl-ddi} and \texttt{ogbl-collab}. The authors of OGB selected Hits@K as the official evaluation metric.  We study the performance of FLAG with GCN and GraphSAGE as backbone on this task. We follow the practice of the baselines to train the models in the full-batch manner. Results are reported in Table \ref{tab:link}. We highlight that FLAG brings a salient boost to both GCN and GraphSAGE on the \texttt{ogbl-ddi} dataset.

{\bf Large-scale Graph Property Prediction.} Table \ref{tab:graph} summarizes the test scores of GCN, GIN, and DeeperGCN on all four OGB graph property prediction datasets. ``Virtual" means the model is augmented with virtual nodes \citep{li2017learning,gilmer2017neural,hu2020open}. As adversarial perturbations are crafted by gradient ascent, it would be unnatural and suboptimal to add noises to discrete input node features \citep{zugner2018adversarial}. We firstly project discrete node features into the continuous space and then adversarially augment the hidden embeddings. On \texttt{ogbg-molhiv}, FLAG yields notable improvements, but when GCN has already been hurt by virtual nodes, FLAG appears to exaggerate the harm. On \texttt{ogbg-molpcba}, GIN-Virtual with FLAG receives an absolute value 1.31\% test AP value increase. Besides node classification and link prediction, FLAG's strong effects on graph classification prove its high versatility.

\section{Ablation Studies and Discussions}\label{sec:ablate}

{\bf Compatibility with graph structure regularizers.} As our augmentation manipulates the input features, it is highly complementary to structure-based regularizers. We validate this point through the experiments below. We mainly focus on two widely-used topological augmentation methods to illustrate \footnote{We also tried DropEdge \citep{rong2019dropedge} but it failed to yield performance gain in the first place.}: (i) Neighbor sampling \citep{hamilton2017inductive} randomly samples neighbors for information aggregation. It not only contributes to GNN scalability but also acts as a structure regularizer. A full-batch GraphSAGE reaches $78.50\pm0.14\%$ test accuracy on \texttt{ogbn-products}, and neighbor sampling alone generalizes the model to $78.70\pm0.36\%$. When FLAG is also used, the test accuracy is increased to $79.36\pm0.57\%$. (ii) Virtual node \citep{gilmer2017neural} adds one synthetic node that connects to all existing nodes. Nearly all the numbers from Table \ref{tab:graph} supports that our method works well with virtual node to generalize GNNs further. Here We highlight one representative group of experiments on \texttt{ogbg-ppa} with GIN as baseline. Vanilla GIN gets $68.92\pm1.00\%$ test accuracy. By adding virtual node alone, it goes to $70.37\pm1.07\%$. When FLAG is further deployed, test accuracy reaches $72.45\pm1.14\%$. 

{\bf Compatibility with batch norm.} Batch norm is appearing more and more frequently in top-performing GNNs. \citet{xie2020adversarial} argued that there was a potential risk, that adversarial examples could distort BN parameters away from natural distribution so to cause the adversarially trained model to fail on clean samples. The authors proposed to use dual batch norms (one for adversaries and the other for clean ones) at training time to better exploit the generalization ability of adversarial augmentations. To test the dual batch norm method on graph data, we run experiments as summarized in Table \ref{tab:bn}. We find that the utilization of dual BN can produce a slight performance gain. As there is growing attention on using batch norms on GNNs, it will be interesting to see how to better synergize adversarial augmentation with batch norms in future research. 
\begin{table}[t] \centering
\caption{Test Accuracy on the \texttt{ogbn-arxiv} dataset with different BN methods.}
\begin{tabular}{lcc}
\Xhline{2\arrayrulewidth}
Method &  GCN  & GraphSAGE \\
\hline\hline  
w/o BN & 71.09\textcolor{gray}{      {$\pm$0.22}} & 69.58\textcolor{gray}{      {$\pm$0.76}}  \\
w/ BN & 71.74\textcolor{gray}{      {$\pm$0.29}}&71.49\textcolor{gray}{      {$\pm$0.27}} \\
w/ BN +FLAG & 72.04\textcolor{gray}{      {$\pm$0.20}}& 72.19\textcolor{gray}{      {$\pm$0.21}}\\
w/ Dual BN +FLAG & \bftab72.11\textcolor{gray}{      {$\pm$0.23}}& \bftab72.21\textcolor{gray}{      {$\pm$0.20}}\\
\Xhline{2\arrayrulewidth}
\end{tabular}

\label{tab:bn}
\end{table}


\begin{table}[t] \centering 

\caption{Test performances on different datasets trained with different adversarial augmentations. Baselines are GAT, GraphSAGE, and GIN respectively. FLAG (fast) means the training epoch number is decreased to make our method trained as fast as the baseline.}

\begin{tabular}{lccc}
\Xhline{2\arrayrulewidth}
   & \texttt{ogbn-products} & \texttt{ogbl-ddi} & \texttt{ogbg-molhiv}\\
 	& Test Acc & Hits@20  & Test ROC-AUC  \\
\hline\hline
Baseline & 79.45\textcolor{gray}{ {$\pm$0.59}} &  53.90\textcolor{gray}{ {$\pm$4.74}} & 75.58\textcolor{gray}{ {$\pm$1.40}} \\
+PGD & 80.96\textcolor{gray}{ {$\pm$0.41}} & 62.02\textcolor{gray}{ {$\pm$6.56}}  & 76.14\textcolor{gray}{ {$\pm$1.62}} \\
+``Free''  & 79.42\textcolor{gray}{ {$\pm$0.84}} & 58.61\textcolor{gray}{ {$\pm$6.0}} &  74.93\textcolor{gray}{ {$\pm$1.29}} \\
\hline
+FLAG  & \bftab81.76\textcolor{gray}{ {$\pm$0.45}} & \bftab63.31\textcolor{gray}{ {$\pm$6.06}} & \bftab76.54\textcolor{gray}{ {$\pm$1.14}} \\
+FLAG (fast)  & 80.64\textcolor{gray}{ {$\pm$0.74}} & - & - \\

\Xhline{2\arrayrulewidth}
\end{tabular}

\label{tab:adv}
\end{table}
\begin{table}[t]  \centering
\caption{Test Accuracy on the \texttt{ogbn-products} dataset.}
\begin{tabular}{lc}
\Xhline{2\arrayrulewidth}
 Backbone	& Test Acc  \\
\hline\hline
GAT w/o dropout & 75.67\textcolor{gray}{ {$\pm$0.27}} \\
GAT w/ dropout & 79.45\textcolor{gray}{ {$\pm$0.59}}\\
GAT w/ dropout +FLAG  & \bftab81.76\textcolor{gray}{ {$\pm$0.45}}\\
\Xhline{2\arrayrulewidth}
\end{tabular}

\label{tab:dp}
\end{table}
\begin{table}[t] \centering
\caption{Test accuracy on \texttt{ogbn-products} with GraphSAGE trained with diverse mini-batch algorithms.}
\centering
\begin{tabular}{lc}
\Xhline{2\arrayrulewidth}
& \texttt{ogbn-products} \\
 Backbone	& Test Acc  \\
\hline\hline
GraphSAGE w/ NS & 78.70\textcolor{gray}{ {$\pm$0.36}} \\
+FLAG & \bftab79.36\textcolor{gray}{ {$\pm$0.57}} \\
\hline
GraphSAGE w/ Cluster & \bftab78.97\textcolor{gray}{ {$\pm$0.33}}\\
+FLAG & 78.60\textcolor{gray}{ {$\pm$0.27}} \\
\hline
GraphSAGE w/ SAINT  & 79.08\textcolor{gray}{ {$\pm$0.24}}\\
+FLAG & \bftab79.60\textcolor{gray}{ {$\pm$0.19}} \\
\Xhline{2\arrayrulewidth}
\end{tabular}

\label{tab:batch}
\end{table}

{\bf Comparison with other robust optimization methods.} Table \ref{tab:adv} shows performances with different adversarial augmentations. For PGD and ``free'', we compute 8 ascent steps for the inner-maximization to make the attack strong enough, while for FLAG we only compute 3 steps. We can see that FLAG outperforms all other methods. We attribute that to the practice of our multi-scale augmentation, which diversifies the scale range of feature perturbations, and helps the model see diverse input features to generalize better, especially on out-distribution samples. Although ``free'' method incorporates diversifying augmentations, but here the benefits are overwhelmed by the suboptimal problem.

\begin{figure*}[th]%
    \centering
    \begin{subfigure}{.3\textwidth}
        \includegraphics[width=\linewidth]{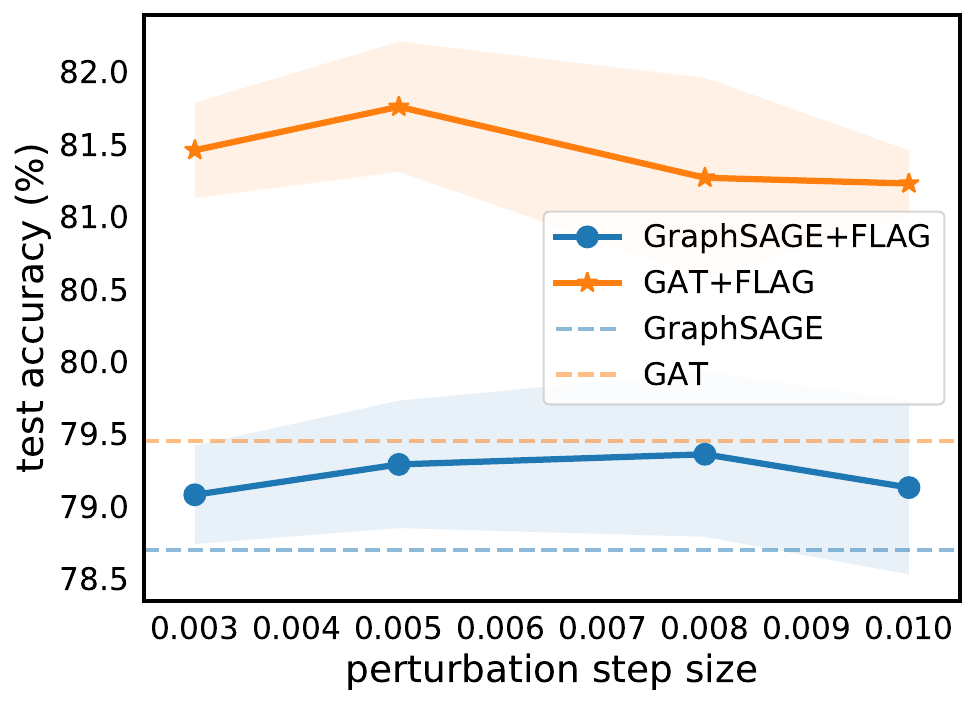} 
        \caption{step size}
        \label{fig:ablate1}
    \end{subfigure}%
    \begin{subfigure}{.3\textwidth}
        \includegraphics[width=\linewidth]{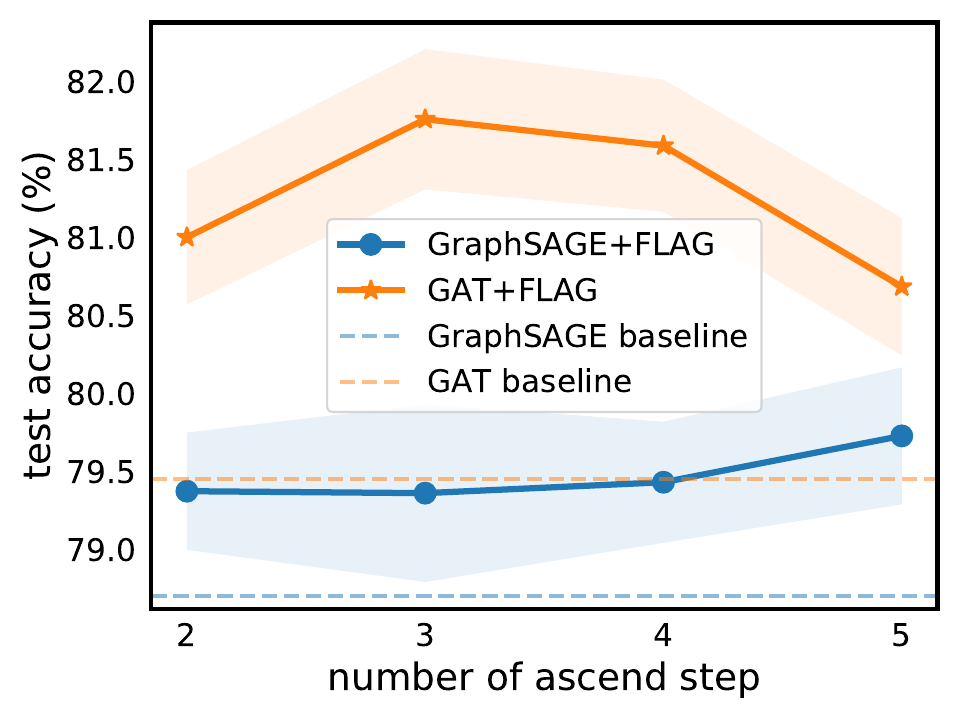}%
        \caption{ascent steps}
        \label{fig:ablate2}
    \end{subfigure}%
    \begin{subfigure}{.3\textwidth}
        \includegraphics[width=\linewidth]{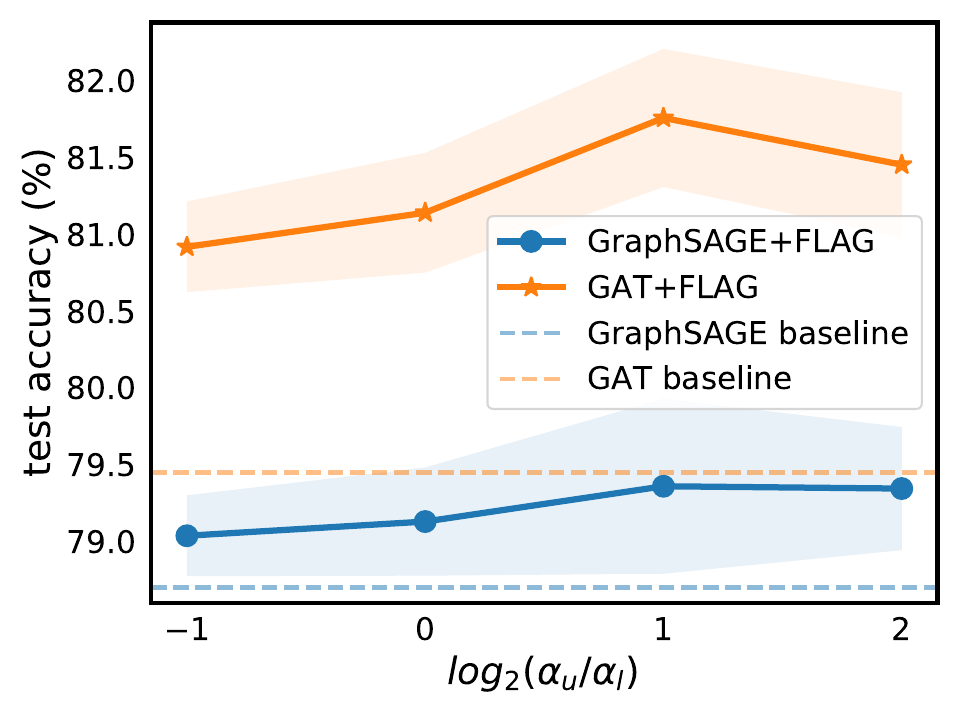}%
        \caption{weighted perturbation}
        \label{fig:biased}
    \end{subfigure}
    \caption{Results of GraphSAGE and GAT on the \texttt{ogbn-products} dataset.}%
\end{figure*}

{\bf Effects of weighted perturbation.} The effects of biased perturbation are reported in Figure \ref{fig:biased}. Generally speaking, when $\text{log}_2(\alpha_u/\alpha_l)>0$, which means that unlabeled nodes receive larger augmentations, the performance gains are more salient. The phenomenon supports our practice of using weighted perturbation to promote multi-scale augmentations. Empirically we find that the benefit of weighted perturbation is more evident on \texttt{ogbn-products} than on \texttt{ogbn-arxiv}. Our understanding is that, \texttt{ogbn-products} is better suited with our practice of labeled vs. unlabeled split because of its high label sparsity compared with \texttt{ogbn-arxiv} (label rate 8\% vs. 54\%). When labeled nodes are more sparse, the neighborhood of labeled nodes will be more overwhelmed by unlabeled ones, where our approximation is more accurate.

\textbf{Hyperparameter sensitivity.} Figure \ref{fig:ablate1} and Figure \ref{fig:ablate2} show the hyperparameter sensitivity of our method. Overall, our method is stable to yield consistent accuracy boost compared with baseline. 


    
    

{\bf Compatibility with mini-batch methods.} Graph mini-batch algorithms are critical to training GNNs on large-scale datasets. We test how different algorithms will work with adversarial data augmentation with GraphSAGE as the backbone. From Table \ref{tab:batch}, we see that neighbor sampling \citep{hamilton2017inductive} and GraphSAINT \citep{zeng2019graphsaint} can all work with FLAG to further boost performance, while Cluster \citep{chiang2019cluster} suffers an accuracy drop.


{\bf Compatibility with dropout.} Dropout is widely used in GNNs. Table \ref{tab:dp} shows that, when trained without dropout, GAT accuracy drops steeply by a large margin. What is more, FLAG can further generalize GNN models together with dropout, similar to the phenomenon of image augmentations. It demonstrates that our method is fully compatible with this domain/model-agnostic regularizer.

{\bf Towards going ``free".} FLAG introduces tractable extra training overhead.  We empirically show that, when we decrease the total number of training epochs to make it as fast as the standard GNN training pipeline, FLAG still brings significant performance gains. Table \ref{tab:adv} shows that FLAG with fewer epochs still generalizes the baseline. Empirically, on a single Nvidia RTX 2080Ti, 100-epoch vanilla GAT takes 88 mins, while FLAG (fast) in Table \ref{tab:adv} takes 91 mins. We note that heuristics like early stopping and cyclic learning rates can further accelerate the adversarial training process \citep{wong2020fast}, so there are abundant opportunities for further research on adversarial augmentation at lower or even no cost.

\begin{figure*}[ht]\small
    \centering
    \begin{subfigure}{.4\textwidth}
        \includegraphics[width=\linewidth]{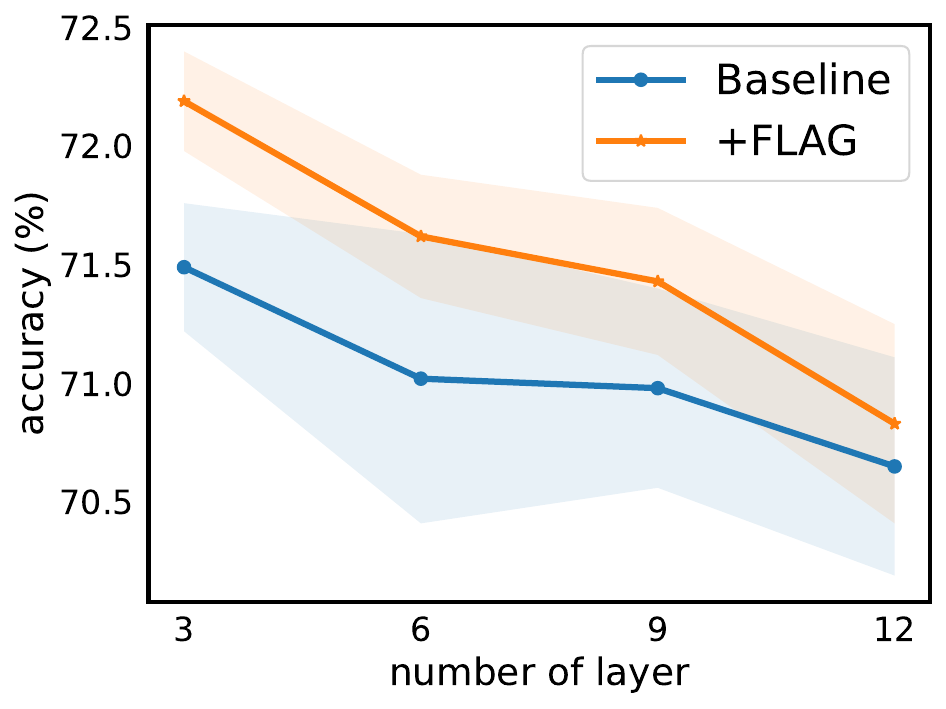} 
        \caption{}
        \label{fig:deep-cora1}
    \end{subfigure}%
    \begin{subfigure}{.4\textwidth}
        \includegraphics[width=\linewidth]{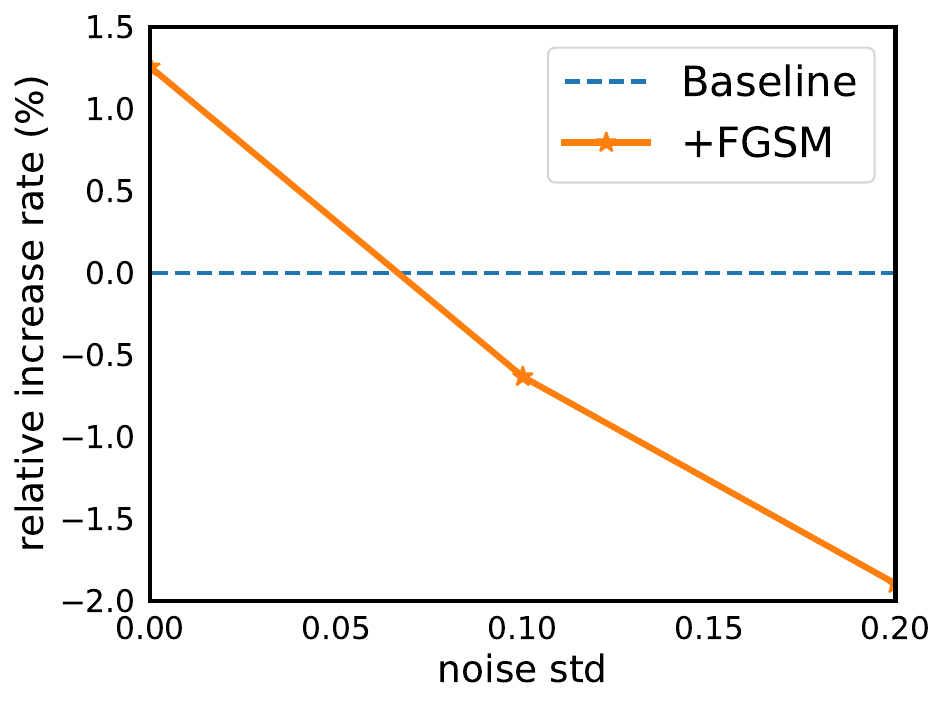}%
        \caption{}
        \label{fig:deep-cora2}
    \end{subfigure}%
    \caption{(a) Test accuracy on \texttt{ogbn-arxiv}; (b) Performance gap on \texttt{Cora}}%
\end{figure*}

{\bf Towards going deep.} Over-smoothing stops GNNs from going deep. FLAG shows its ability to boost both shallow and deep baselines, e.g., GCN and DeeperGCN. We carefully examine FLAG's effects on generalization when a GNN goes progressively deeper in Figure \ref{fig:deep-cora1}. The experiments are conducted on \texttt{ogbn-arxiv} with GraphSAGE as the backbone, where a consistent improvement is evident.


{\bf What if there's no node feature?} One natural question can be raised: what if no input node features are provided? \texttt{ogbn-proteins} is a dataset without input node features. \citet{hu2020open} proposed to average incoming edge features to obtain initial node features, while \cite{li2020deepergcn} used summation and achieved competitive results. Note that the GCN and GraphSAGE baselines in Table \ref{tab:node-main} use the ``mean" node features as input and suffer an accuracy drop with FLAG; DeeperGCN leverages the ``sum" and gets further improved. Interestingly, when DeeperGCN is trained with ``mean" node features, it receives high invariance, so that even large magnitude perturbations will not change its result. The diverse behavior of adversarial augmentation implies the importance of node feature construction method selection.

{\bf Where Does the Boost Come from?} It is now widely believed that model robustness appears to be at odds with clean accuracy. Despite the recent proliferation of literature in using adversarial data augmentation to promote standard performance, it is still unsettled where the boost or detriment of adversarial training comes from. Like one-hot word embeddings for language models, input node features usually come from discrete spaces, e.g., the bag-of-words binary features in \texttt{ogbn-products}. We conjecture that the diverse effects of adversarial training in different domains stem from differences in the input data distribution rather than model architectures. To ground our claim, we have the following observations. 

\textit{Observation 1:} We utilize FLAG to augment MLPs (an architecture where adversarial training has adverse effects in the image domain), and successfully boost generalization. FLAG directly improves the test accuracy from $61.06\pm0.08\%$ to $62.41\pm0.16\%$ on \texttt{ogbn-product}, and from $55.50\pm0.23\%$ to $56.02\pm0.19\%$ on \texttt{ogbn-arxiv}. 

\textit{Observation 2:} In general, adversarial training hurts the clean accuracy in image classification, but \citet{tsipras2018robustness} showed that CNNs could benefit from adversarial augmentations on MNIST, where the pixel values are closer to discrete distribution than other more natural image datasets. 

\textit{Observation 3:} To illustrate, we provide a simple example on the \texttt{Cora}~\citep{getoor2005link} dataset. To simplify the scenario, we choose FGSM to craft adversarial augmentations for a GCN. By adding Gaussian noise with standard deviation $\sigma$, we simulate node features drawn from a continuous distribution. The result is summarized in Figure \ref{fig:deep-cora2}. When $\sigma=0$, the discrete distribution of node features persists. At this moment, a GCN with adversarial augmentation outperforms the non-augmented model. With increased noise level $\sigma$, the features are continuously distributed with large support and FGSM starts to harm the clean accuracy, which validates our conjecture. All these observations support our conjecture that data distribution has more to do with the effect of adversarial augmentation, while the lack of theoretical justification is a limitation of our analysis.

\section{Conclusion}

We propose FLAG, a simple,  scalable, and general data augmentation method for better GNN generalization.  Like widely-used image augmentations, FLAG can be easily incorporated into any GNN training pipeline. FLAG yields improvements over a range of GNN baselines. Besides extensive experiments, we also provide conceptual analysis to validate adversarial augmentation's different behavior on varied data types. The effects of adversarial augmentation on generalization are still not entirely understood, and we think this is a fertile space for future exploration.

\section{Acknowledgement}
Kezhi Kong and Tom Goldstein were supported by DARPA GARD, Office of Naval Research, AFOSR MURI program, the DARPA Young Faculty Award, and the National Science Foundation Division of Mathematical Sciences. Additional support was provided by Capital One Bank and JP Morgan Chase. Guohao Li and Bernard Ghanem were supported by the King Abdullah University of Science and Technology (KAUST) Office of Sponsored Research through the Visual Computing Center (VCC) funding.

\clearpage
\newpage
\bibliography{reference}

\clearpage
\newpage
\appendix
\section{Appendix}
\label{appendix}

\begin{figure}[h]%
    \centering
    \includegraphics[width=\linewidth]{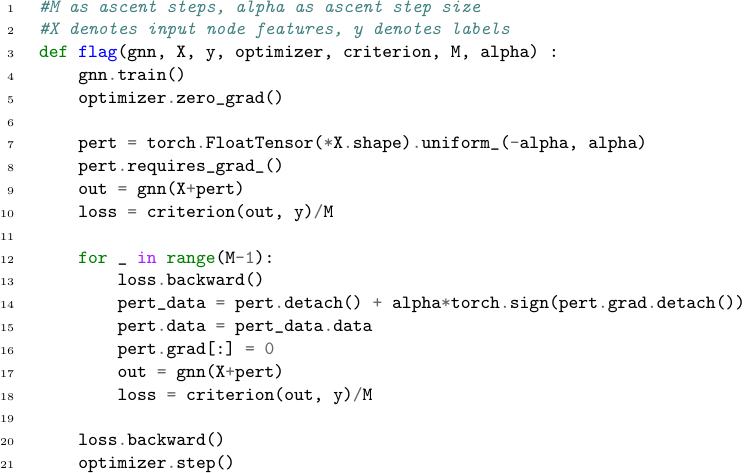} %
    \caption{An abstract PyTorch Implementation of our method.}%
    \label{fig:pseudo}%
\end{figure}

We summarize implementation details and selected hyperparameters in this section. Note that for ALL of our method, we fix the ascent step number M to 3 for simplicity. We leave more thorough step number search for future research. Experiments are done on hardware with Intel(R) Xeon(R) Silver 4216 CPU @ 2.10GHz, and 128GB of RAM. If without mentioning, we use a single GeForce RTX 2080 Ti (11GB GPU memory). To highlight, for fair comparisons, we do not modify model architectures nor optimizing algorithms.

\subsection{Node Classification}

\subsubsection{ogbn-products}

\textbf{MLP:} perturbation step size $\alpha$=2e-02, only labeled nodes are used in the training phase.

\textbf{GraphSAGE:} labeled perturbation step size $\alpha_l$=8e-03, $\alpha_u/\alpha_l$=2, and neighbor sampling is used for scalable training.

\textbf{GAT:} labeled perturbation step size $\alpha_l$=5e-03, $\alpha_u/\alpha_l$=2, and neighbor sampling is used for scalable training.

\textbf{DeeperGCN:} labeled perturbation step size $\alpha_l$=5e-03, $\alpha_u/\alpha_l$=2, and the model is trained on NVIDIA Tesla V100 (32GB GPU memory).

\subsubsection{ogbn-proteins}

\textbf{GCN:} labeled perturbation step size $\alpha_l$=1e-03, $\alpha_u/\alpha_l$=1, and the model is trained in the full-batch manner.

\textbf{GraphSAGE:} labeled perturbation step size $\alpha_l$=1e-03, $\alpha_u/\alpha_l$=1, and the model is trained in the full-batch manner.

\textbf{DeeperGCN:} labeled perturbation step size $\alpha_l$=8e-03, $\alpha_u/\alpha_l$=1, and the model is trained on NVIDIA Tesla V100 (32GB GPU memory).

\subsubsection{ogbn-arxiv}

\textbf{MLP:} perturbation step size $\alpha$=2e-03, only labeled nodes are used in the training phase.

\textbf{GCN:} labeled perturbation step size $\alpha_l$=1e-03, $\alpha_u/\alpha_l$=1, and the model is trained in the full-batch manner.

\textbf{GraphSAGE:} labeled perturbation step size $\alpha_l$=1e-03, $\alpha_u/\alpha_l$=1, and the model is trained in the full-batch manner.

\textbf{GAT:} labeled perturbation step size $\alpha_l$=1e-03, $\alpha_u/\alpha_l$=2, and the model is trained in the full-batch manner.

\textbf{DeeperGCN:} labeled perturbation step size $\alpha_l$=8e-03, $\alpha_u/\alpha_l$=1, and the model is trained on NVIDIA Tesla V100 (32GB GPU memory).

\subsubsection{ogbn-mag}

\textbf{R-GCN:} labeled perturbation step size $\alpha_l$=1e-04, $\alpha_u/\alpha_l$=1, and the model is trained with neighbor sampling for scalability.

\subsection{Link Prediction}

\subsubsection{ogbl-ddi}

\textbf{GCN} perturbation step size $\alpha$=3e-03, and \textbf{GraphSAGE} perturbation step size $\alpha_l$=3e-03. Models are both trained in the full-batch manner. During each gradient ascent loop negative edges are resampled for computing negative losses. 

\subsubsection{ogbl-collab}

\textbf{GCN} perturbation step size $\alpha$=3e-03, and \textbf{GraphSAGE} perturbation step size $\alpha_l$=3e-03. Models are both trained in the full-batch manner. During each gradient ascent loop negative edges are resampled for computing negative losses. 

\subsection{Graph Classification}

\subsubsection{ogbg-molhiv}

\textbf{GCN:} perturbation step size $\alpha$=1e-02, when virtual node is added we use a smaller $\alpha$=1e-03.

\textbf{GIN:} perturbation step size $\alpha$=5e-03, when virtual node is added we use a smaller $\alpha$=1e-03.

\textbf{DeeperGCN:} perturbation step size $\alpha$=1e-02, and the model is trained on NVIDIA Tesla V100 (32GB GPU memory).

\subsubsection{ogbg-molpcba}

\textbf{GCN:} perturbation step size $\alpha$=8e-03 for both the vanilla model and the one augmented by virtual node.

\textbf{GIN:} perturbation step size $\alpha$=8e-03 for both the vanilla model and the one augmented by virtual node.

\textbf{DeeperGCN:} perturbation step size $\alpha$=8e-03 with virtual node added, and the model is trained on NVIDIA Tesla V100 (32GB GPU memory).

\subsubsection{ogbg-ppa}

\textbf{GCN:} perturbation step size $\alpha$=2e-03, when virtual node is added we use a larger $\alpha$=5e-03.

\textbf{GIN:} perturbation step size $\alpha$=8e-03, when virtual node is added we use a smaller $\alpha$=5e-03.

\textbf{DeeperGCN:} perturbation step size $\alpha$=8e-03, and the model is trained on NVIDIA Tesla V100 (32GB GPU memory).

\subsubsection{ogbg-code}

\textbf{GCN:} perturbation step size $\alpha$=8e-03 for both the vanilla model and the one augmented by virtual node.

\textbf{GIN:} perturbation step size $\alpha$=8e-03 for both the vanilla model and the one augmented by virtual node.

\textbf{DeeperGCN:} perturbation step size $\alpha$=8e-03 with virtual node added, and the model is trained on NVIDIA Tesla V100 (32GB GPU memory).

\section{Dataset Details}

Table \ref{tab:nodedata}, Table \ref{tab:linkdata}, and Table \ref{tab:graphdata} summarize the dataset statistics for node classification, link prediction, and graph classification respectively. Datasets can be downloaded at \url{https://ogb.stanford.edu/}.

\section{Loss Landscape Visualization}

\begin{figure*}[ht]%
    \centering
    \begin{subfigure}{.24\textwidth}
        \includegraphics[width=\linewidth]{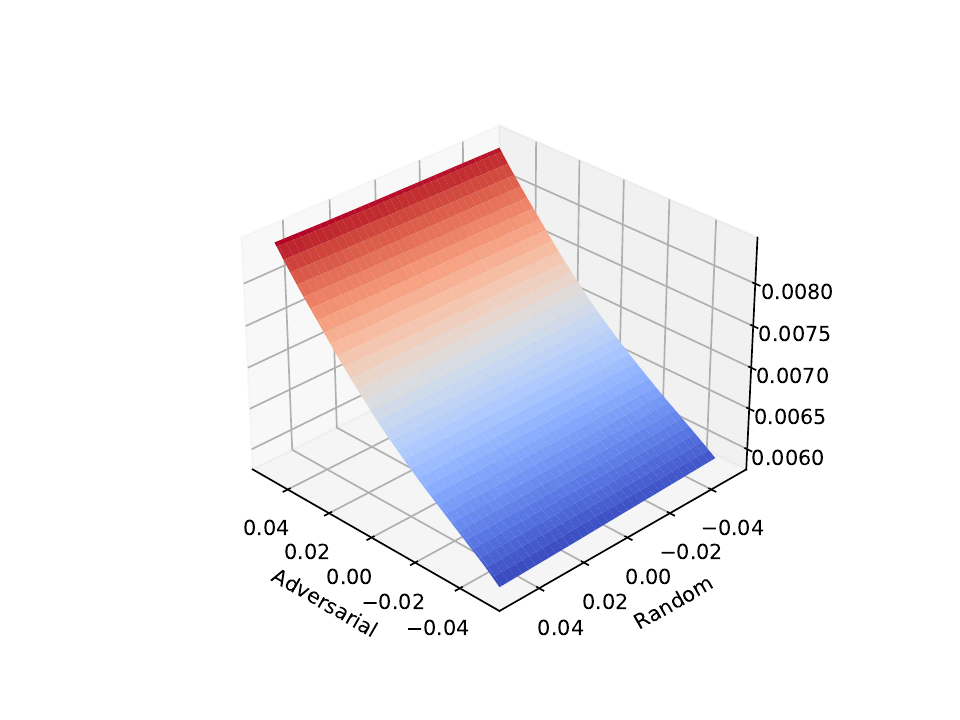} 
        \caption{FLAG adv-random}
    \end{subfigure}%
    \begin{subfigure}{.24\textwidth}
        \includegraphics[width=\linewidth]{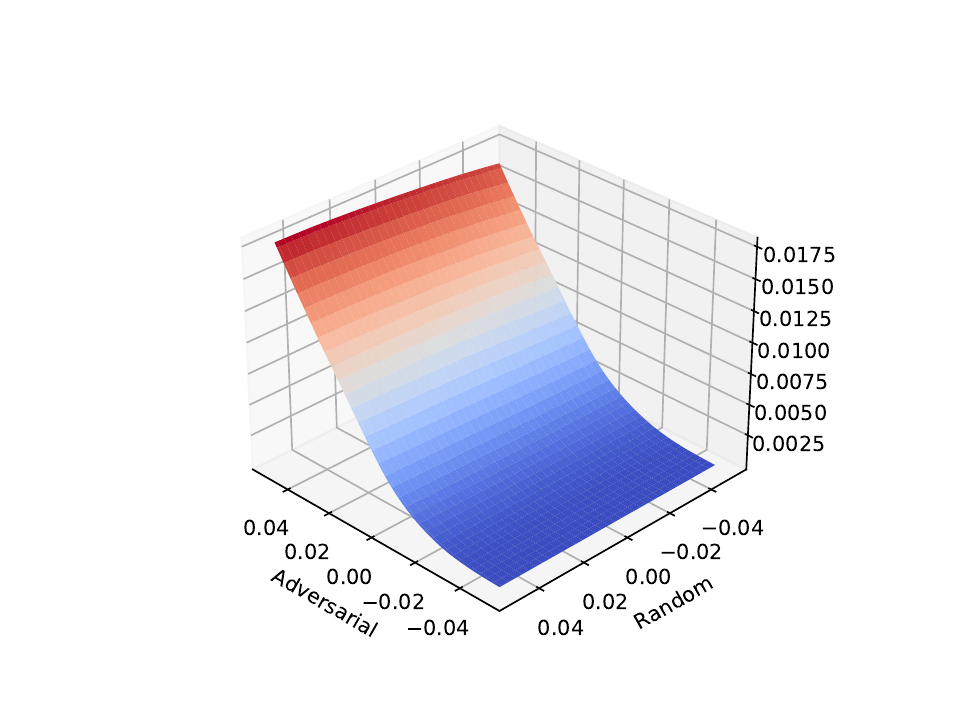}%
        \caption{Vanilla adv-random}
    \end{subfigure}%
    \begin{subfigure}{.24\textwidth}
        \includegraphics[width=\linewidth]{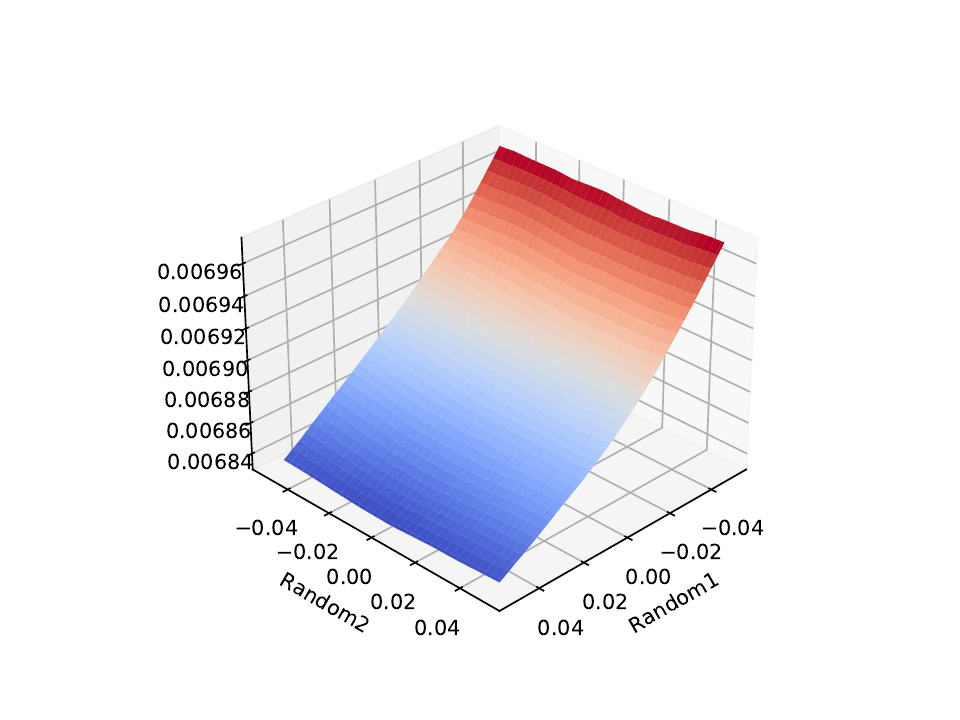}%
        \caption{FLAG random-random}
    \end{subfigure}
        \begin{subfigure}{.24\textwidth}
        \includegraphics[width=\linewidth]{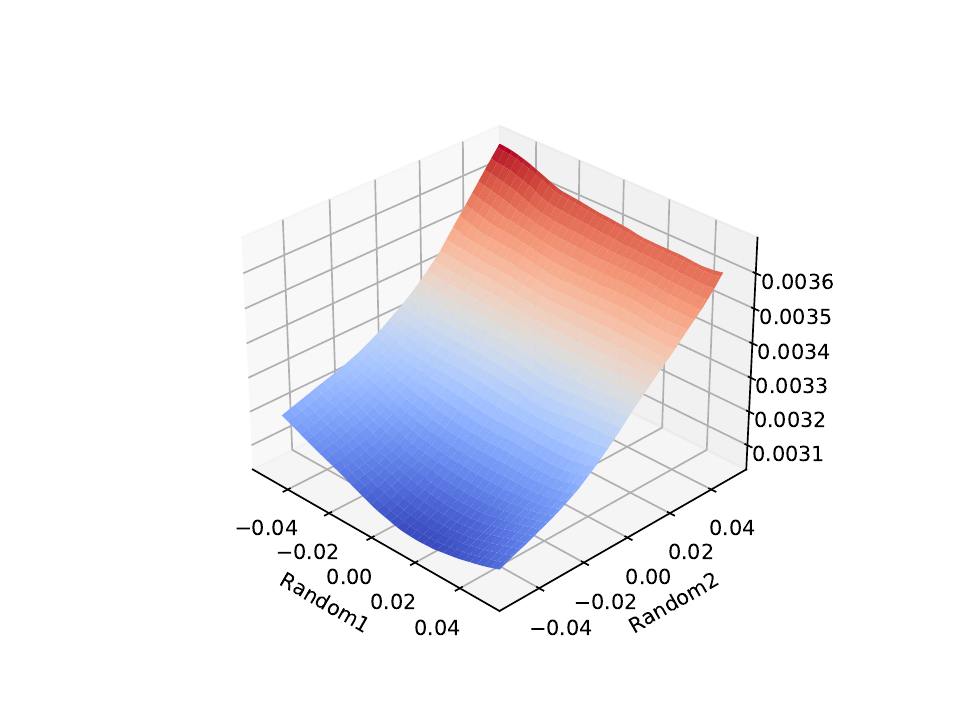}%
        \caption{Vanilla random-random}
    \end{subfigure}
    \caption{Loss landscape visualization. The test is conducted on one random validation graph from  \texttt{ogbg-molhiv}. Two models are GIN trained with FLAG and a vanilla GIN. (a) and (b) projects loss onto a random direction and the other adversarial direction, while (c) and (d) use two random directions.}
    \label{fig:vis}
\end{figure*}

Figure \ref{fig:vis} shows the loss landscape of GIN model. We can see that our method further regularizes the loss landscape.

\begin{table*}
\centering
\scalebox{0.8}{
\begin{tabular}{ccccccc}
\Xhline{2\arrayrulewidth}

 Name	& \#Nodes & \#Edges & \#Tasks & Train/Val/Test & Task Type & Metric \\
\hline
ogbn-products & 2,449,029 & 61,859,140 & 1 & 8/2/90 & Multi-class classification & Accuracy \\
ogbn-proteins & 132,534 & 39,561,252 & 112 & 65/16/19 & Binary classification & ROC-AUC \\
ogbn-arxiv & 169,343 & 1,166,243 & 1 &  54/18/28 & Multi-class classification	& Accuracy \\
ogbn-mag & 1,939,743 & 21,111,007 & 1 &  85/9/6 & Multi-class classification & Accuracy \\
Cora &  2485 &  5069 & 1 & no official split$\ast$ & Multi-class classification & Accuracy \\

\Xhline{2\arrayrulewidth}
\end{tabular}
}
\caption{Node classification datasets statistics. $\ast$ denotes we follow the split of \citet{kipf2016semi}.}

\label{tab:nodedata}
\end{table*}

\begin{table*}
\centering
\begin{tabular}{cccccc}
\Xhline{2\arrayrulewidth}

 Name	& \#Nodes & \#Edges & Train/Val/Test  & Task Type & Metric \\
\hline
ogbl-ddi & 4,267	 & 1,334,889 &  80/10/10 & Link prediction & Hits@20 \\

ogbl-collab & 235,868 & 1,285,465 &  92/4/4 & Link prediction & Hits@50 \\

\Xhline{2\arrayrulewidth}
\end{tabular}
\caption{Link prediction datasets statistics.}

\label{tab:linkdata}
\end{table*}

\begin{table*}[th]
\centering
\scalebox{0.8}{
\begin{tabular}{cccccccc}
\Xhline{2\arrayrulewidth}

 Name	& \#Graphs & Avg \#Nodes  & Avg \#Edges & \#Tasks & Train/Val/Test  & Task Type & Metric \\
\hline
ogbg-molhiv & 41,127 & 25.5 & 27.5 & 1 &  80/10/10 & Binary classification & ROC-AUC \\

ogbg-molpcba & 437,929 & 26.0  & 28.1 & 128  & 80/10/10 & Binary classification	& AP \\

ogbg-ppa & 158,100 & 243.4	& 2,266.1 & 1 &  49/29/22 & Multi-class classification	& Accuracy \\

ogbg-code & 452,741	& 125.2	 & 124.2 & 1 & 90/5/5 & Sub-token prediction	& F1 score \\

\Xhline{2\arrayrulewidth}
\end{tabular}
}
\caption{Graph classification datasets statistics.}

\label{tab:graphdata}
\end{table*}

\end{document}